\documentclass[10pt,twocolumn,letterpaper]{article}
\usepackage{btas}

\usepackage{times}
\usepackage{epsfig}
\usepackage{microtype}
\usepackage{graphicx}		
\usepackage{subcaption}		
\usepackage{booktabs} 		
\usepackage{amsfonts}       
\usepackage{amsmath}
\usepackage{amssymb}
\usepackage{nicefrac}       
\usepackage{bchart}
\usepackage{multirow}
\usepackage{algorithm}
\usepackage{algorithmic}
\usepackage[absolute]{textpos}
\usepackage{balance}

\usepackage[pagebackref=true,breaklinks=true,letterpaper=true,colorlinks,bookmarks=false]{hyperref}

\btasfinalcopy 

\TPMargin{5pt}
\newcommand{\copyrightstatement}{
    \begin{textblock}{0.84}(0.08,0.93)   
    \noindent
    \footnotesize
    978-1-5386-7180-1/18/\$31.00 \copyright 2018 IEEE
    \end{textblock}
}

\ifbtasfinal\fi

\begin{document}
\copyrightstatement

\title{DeepMasterPrints: Generating MasterPrints for Dictionary Attacks via Latent Variable Evolution\thanks{This work was supported by the United States National Science Foundation under Grant 1618750 and Grant 1617466. This project also benefited from GPUs donated by NVIDIA.}}

\author{Philip Bontrager\\
New York University Tandon\\
{\tt\small philipjb@nyu.edu}
\and
Aditi Roy\\
New York University Tandon\\
{\tt\small ar3824@nyu.edu}
\and
Julian Togelius\\
New York University Tandon\\
{\tt\small julian@togelius.com}
\and
Nasir Memon\\
New York University Tandon\\
{\tt\small memon@nyu.edu}
\and
Arun Ross\\
Michigan State University\\
{\tt\small rossarun@cse.msu.edu}
}


\maketitle
\thispagestyle{empty}

\begin{abstract}

Recent research has demonstrated the vulnerability of fingerprint recognition systems to dictionary attacks based on MasterPrints. MasterPrints are real or synthetic fingerprints that can fortuitously match with a large number of fingerprints thereby undermining the security afforded by fingerprint systems. Previous work by Roy et al. generated synthetic MasterPrints at the feature-level. In this work we generate complete image-level MasterPrints known as DeepMasterPrints, whose attack accuracy is found to be much superior than that of previous methods. The proposed method, referred to as Latent Variable Evolution, is based on training a Generative Adversarial Network on a set of real fingerprint images. Stochastic search in the form of the Covariance Matrix Adaptation Evolution Strategy is then used to search for latent input variables to the generator network that can maximize the number of impostor matches as assessed by a fingerprint recognizer. Experiments convey the efficacy of the proposed method in generating DeepMasterPrints. The underlying method is likely to have broad applications in fingerprint security as well as fingerprint synthesis.

\end{abstract}

\section{Introduction}

Fingerprints are increasingly being 
used to verify the identity of an individual in a  variety of applications ranging from unlocking doors to securing smartphones to authorizing payments. 
In some  applications such as smartphones,  the fingerprint sensor is small in size for ergonomic reasons \cite{han2014fingerprint} and, therefore, these sensors obtain only partial images of a user's fingerprint. Since small portions of a fingerprint are not as distinctive as the full fingerprint, the chances of a partial fingerprint (from one finger) being incorrectly matched with another partial fingerprint (from a different finger) are higher. This observation was exploited by Roy \etal \cite{masterprint}, who introduced the notion of {\em MasterPrints}. MasterPrints are a set of real or synthetic fingerprints that can fortuitously match with a large number of other fingerprints. Therefore, they can be used by an adversary to launch a dictionary attack against a specific subject that can compromise the security of a fingerprint-based recognition system. This means, it is possible to  ``spoof" the fingerprints of a subject without actually gaining any information about the subject's fingerprint.  




Roy \etal \cite{masterprint} demonstrated that MasterPrints can either be obtained from {\em real} fingerprint images or can be {\em synthesized} using a hill-climbing procedure. The synthetic MasterPrints were generated at the ``template level'' by modifying the minutiae points in a fingerprint template \cite{masterprint,RoyMasterPrint_ICB2018}. The methods in \cite{masterprint,RoyMasterPrint_ICB2018} did not generate images. However, to launch a spoof-attack in practice, it is necessary to construct MasterPrints at the ``image level" which can then be transferred to a  physical artifact. 
This observation motivated us to find a method for generating DeepMasterPrints - images that are visually similar to natural fingerprint images.

To design DeepMasterPrints, there needs to be a way to feasibly search the space of fingerprint images. Since not all fingerprint systems use minutiae \cite{ross2003hybrid,RossCorrelation_ECCVW2004}, it is advantageous if minutiae information is not explicitly used during the design process (unlike \cite{masterprint,RoyMasterPrint_ICB2018}). Now neural networks can be used to generate synthetic fingerprint images. In particular, Generative Adversarial Networks (GANs) have shown great promise in generating images that reproduce a particular style or domain~\cite{CaoJainFPSynthesisICB2018, og_gan, nipsTutorial}. {\bf However, their standard design is not controllable.} In other words, they do not allow the generator to target additional constraints and objectives beyond reproducing the style of the training data. For a DeepMasterPrint, we need to create a synthetic fingerprint image that can fool a fingerprint matcher. The matcher should not only realize that the image is a fingerprint (visual realism), but should also match that fingerprint image to many different identities. Therefore, a generator network has to be combined with a {\em method} of searching for DeepMasterPrints.

In this paper, we present a method for creating DeepMasterPrints. This technique uses a neural network to learn to generate images of fingerprints. It then uses evolutionary optimization to search the latent variable space of the neural network for a DeepMasterPrint. The Covariance Matrix Adaptation Evolution Strategy (CMA-ES) is used to search the input space of the trained neural network for the ideal fingerprint image. This unique combination of evolutionary optimization and generative neural networks allows the neural network to constrain the search space while the evolutionary algorithm handles the discrete fitness function.

This is the first work that creates a {\em synthetic} Masterprint at the {\em image-level} thereby further reinforcing the danger of utilizing small-sized sensors with limited resolution in fingerprint applications. This work directly shows how to execute this exploit and is able to spoof 23\% of the subjects in the dataset at a 0.1\% false match rate. At a 1\% false match rate, the generated DeepMasterPrints can spoof 77\% of the subjects in the dataset. 


\section{Background}

\subsection{Dictionary attack using synthetic MasterPrints}

Research in assessing vulnerabilities in a fingerprint recognition system is a constant arms race between fixing vulnerabilities and discovering new ones \cite{RathaBioPrivacy01}. It is important for researchers to probe for new vulnerabilities so that loopholes can be fixed \cite{MarascoAntiSpoofing_ACM2015}. Sometimes it is not just necessary to prove that a vulnerability exists, but to show how an attack can actually be executed \cite{galbally2010evaluation}. This has two important consequences for researchers designing secure systems: (a) it allows them to evaluate the immediate risk of this threat, and (b) it gives them a concrete attack vector to protect against. Research around the vulnerability to MasterPrints is important for these reasons.


As stated earlier, a MasterPrint is a real or synthesized fingerprint that can be used to impersonate multiple identities. This type of attack does not require knowledge of a specific individual's fingerprint sample; instead, the attack can be launched against anonymous subjects with some probability of success \cite{wolfattack}. The attack itself exploits the vulnerability of small fingerprint sensors that only image a portion of the fingerprint. Such sensors may not scan the entire fingerprint and, therefore, only partial prints are available. Since it would be impractical to require the user to place their finger the exact same way every time, these systems normally take multiple readings from the same finger during enrollment. When a partial fingerprint is presented to the system during verification, it is compared against all the partial enrolled prints  corresponding to the subject. If a subject has $n$ fingers in the system and there are $k$ partial prints saved per fingerprint, then there are $n \times k$ opportunities for a match and the input image only needs to match one of them to be declared a success. Such a setup is common on consumer mobile devices hosting small fingerprint sensors.

Roy et al. \cite{masterprint} showed that MasterPrints could be extracted from real fingerprints or could be synthetically generated. In the latter case, the authors generated synthetic {\em minutiae} templates. Minutiae points in a fingerprint correspond to ridge endings and ridge bifurcations. Each minutia point is represented as a 3-tupled value, ($x,y,\theta$), where ($x,y$) denotes the location of the minutia and $\theta$ denotes the local orientation of the ridge on which the minutia is located. The authors used a hill-climbing algorithm to iteratively modify and synthesize a minutiae template that could be employed as a MasterPrint. The objective function for the hill-climbing procedure was the number of distinct fingerprint templates in a training database that were successfully matched with the synthetic template. Their approach, however, has two distinct disadvantages: Firstly, it does not generate an image. Although one could potentially reconstruct an image from the template \cite{RossReconstruction07}, these images have a very synthetic look and could likely be detected.
Secondly, their approach is applicable primarily to minutiae-based matchers. Matchers that utilize other information (e.g., local ridge frequency and orientation) may not be vulnerable to such synthetic minutiae templates.   
 
In this work, we directly generate {\em images} instead of  minutiae templates. One advantage of generating images instead of templates, is that it is theoretically possible to design DeepMasterPrints for any fingerprint system that accepts images \cite{ios}.
Further, the attack can potentially be launched at the sensor level by transferring the images to a spoof artifact. 

\subsection{Image generation}

Recently, there have been rapid advancements in synthetic image generation by way of neural networks. Some of the most popular methods for image generation are Fully Visible Belief Networks (FVBN), Variational Autoencoders (VAE), and Generative Adversarial Networks (GAN) \cite{nipsTutorial}. FVBNs such as PixelRNN produce one pixel at a time, similar to text generation, but can often be noisy in their output. VAEs, on the other hand, tend to produce very smooth outputs. Current GAN methods are perceived to produce results with fewer artifacts than FVBNs and sharper images than VAEs \cite{nipsTutorial}. In the end, any of these methods could be used in this work as long as they generate good quality fingerprint images. 

GANs learn to generate images in an unsupervised fashion. There are two parts to a GAN: a generator and a discriminator. The generator is typically a neural network that inputs random noise and outputs an image. The discriminator is also typically a neural network, which inputs an image and classifies it as being `real' or `generated'. To ensure that the generator produces images within the domain of the sample images, training happens in three steps: (a) Provide real images to the discriminator. Train the discriminator to classify them as real. (b) Provide generated images to the discriminator. Train the generator to classify them as generated. (c) Provide the generator with the discriminator's gradients. Train the generator to produce images that are classified as real.




This process is repeated until the network converges on an approximation of the distribution of the real data. 

A major difficulty during training is keeping the two networks balanced so one does not become significantly better than the other. Much work, since the invention of GANs, has focused on stabilizing the training process; two popular approaches are the Wasserstein GAN (WGAN) and WGAN with gradient penalty \cite{wgan, wgangp}. In standard GAN training, the discriminator classifies the input as being either `real' or `generated'. The difference between the real data distribution and the generated data distribution is then measured using the Jensen-Shannon divergence (JS) metric \cite{wgan}. This metric does not provide a gradient everywhere for the generator and, therefore, requires the discriminator and generator to be closely matched. This, in turn, makes training unstable. WGAN, instead, uses an approximation of the Wasserstein distance function to measure the difference between the real and generated distributions \cite{wgan}. Since it is differentiable everywhere, it provides meaningful gradients for the generator. The two networks do {\em not} have to be well balanced and so the discriminator can be better trained preventing mode collapse.

While we had success with WGAN in this work, in principle, any GAN algorithm could have been used. A recent study tested a number of GAN algorithms and found that with enough parameter tuning there was not a significant difference between them \cite{equal_gans}. If larger images are needed, a recent work shows that progressively growing GANs produces good results \cite{progressive_gan}.

\subsection{Evolutionary Optimization}

Optimization via evolutionary strategies has been used in AI for a long time. Evolutionary computation is a family of versatile optimization techniques that only need a method for representing and comparing solutions to find an optimal solution. The basic algorithm starts with a random sample of solutions, or members, from a population of all represented solutions. The algorithm then evaluates the sample and ranks each member. The best members are then varied to get a new sample of potentially superior solutions. This process is repeated until convergence or some other constraint is met.
Evolution is particularly suited to instances where the evaluation mechanism is a black box and only the final evaluation of each sample is available.

The Covariance Matrix Adaption Evolutionary Strategy (CMA-ES) is a robust approach that has been shown to work on non-linear and non-convex fitness domains \cite{cmaes}. CMA-ES samples its population from a multivariate normal distribution. Since each solution is represented as a combination of variables, CMA-ES maintains a covariance matrix that tracks how each variable affects fitness. In each generation, it creates a new sample based on the information in the covariance matrix. If variables A and B are shown to be highly correlated in solutions with a high fitness, then it's highly likely that the new sampled members will have A and B correlated. This matrix is updated based on the fitness of the new sample allowing the algorithm to learn the distribution of successful samples. The model it learns is an approximation of a second-order model of the fitness function \cite{cmaes}. This makes CMA-ES a powerful strategy when optimizing for difficult real-valued domains.

Evolutionary methods have been used with neural networks for a long time. This has primarily occurred through neuroevolution, where evolution is used to evolve the weights and, sometimes, the topology of a neural network \cite{evolvenn}. Recently, researchers have shown that neuroevolution can be used on deep neural networks and can even compete against reinforcement learning algorithms \cite{deep_evolve, verydeep_evolve}. Our work does not involve evolving the weights of the neural networks. Instead, in our work, the neural networks are trained separately using the gradient descent algorithm, but evolution is applied to the network inputs. A similar approach has recently been proposed for an interactive evolution system \cite{deepIE}; the difference there is that human aesthetic preference is used as the fitness function.

\section{Proposed Methods}

The ideal system for generating a DeepMasterPrint would be able to (a) generate every possible image, (b) test each image on all fingerprint matchers in existence, and (c) choose the image that successfully matches against the most number of distinct fingerprints pertaining to a large number of identities. Since it is infeasible to have access to every fingerprint matcher, it is necessary to derive a DeepMasterPrint based on a sample of identities and matchers, and have it generalize. Limiting the images to just images of fingerprints helps in generalization. Our scaled back ideal system is able to generate any fingerprint image and search over a sample of identities and matchers to find an ideal solution. This approach not only generates an image, but it also has the potential to find a more effective solution than previous approaches. To implement this approach, we developed a new technique called Latent Variable Evolution.

There are two parts to Latent Variable Evolution (LVE); (1) train a neural network to generate images of fingerprints, and (2) search over the latent variables of the network (the input vector to the generator network) for a fingerprint that results in the best DeepMasterPrint, i.e., a fingerprint image that matches with a large number of other fingerprint images. To train an image generator, we use the WGAN method described earlier and then use CMA-ES to evolve the fingerprint. The method is tested on two different fingerprint datasets and with several different matchers.

\subsection{Fingerprint Generator}

\begin{figure}
	\centerline{\includegraphics[width=.7\linewidth]{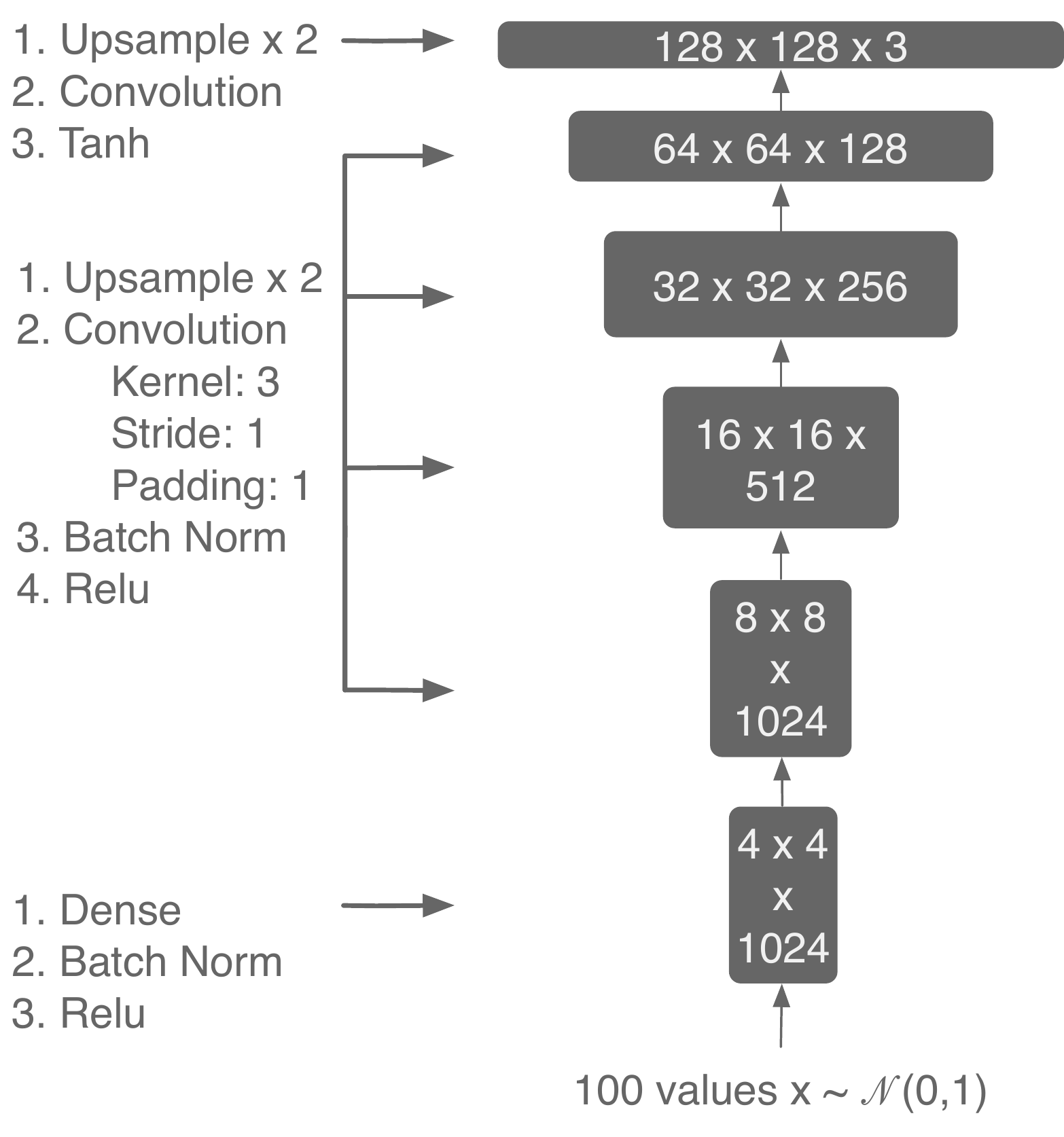}}
  	\caption{Generator Network Architecture. The discriminator is the inverse of this model with subsampling instead of upsampling and LeakyReLu activation functions.}
  	\label{fig:network}
\end{figure}

In this work we train two generator networks, both using the WGAN algorithm. The networks are modeled after deep convolutional GAN and defined in Figure\ref{fig:network} \cite{dcgan}. One network is trained on a dataset of fingerprints scanned with a capacitive sensor, and the other on a dataset of inked and rolled fingerprints. The networks are trained adversarially with a Wasserstein loss function and RMSProp with a learning rate of 0.00005 \cite{wgan}. The generators are trained using the minibatch gradient descent scheme. Each batch samples 64 images and 64 latent variable vectors. We trained each generator for 120,000 updates, with the discriminator being trained 5 times between each generator update. Using deconvolutions for the generator resulted in blocky artifacts therefore we switched to upsampling with convolutions.

It should be noted that there has been some research conducted into analyzing how much of the data distribution a GAN actually learns \cite{birthdayParadox}. Arora and Zhang tested a few GAN architectures for diversity, and none of the GANs that they tested were able to model the entire data distribution well. In the light of this, we acknowledge that WGAN will only model part of the image distribution and our technique will not be able to produce every type of fingerprint.

\subsection{Searching the Space of Latent Variables}
\begin{figure*}
	\includegraphics[width=1\linewidth]{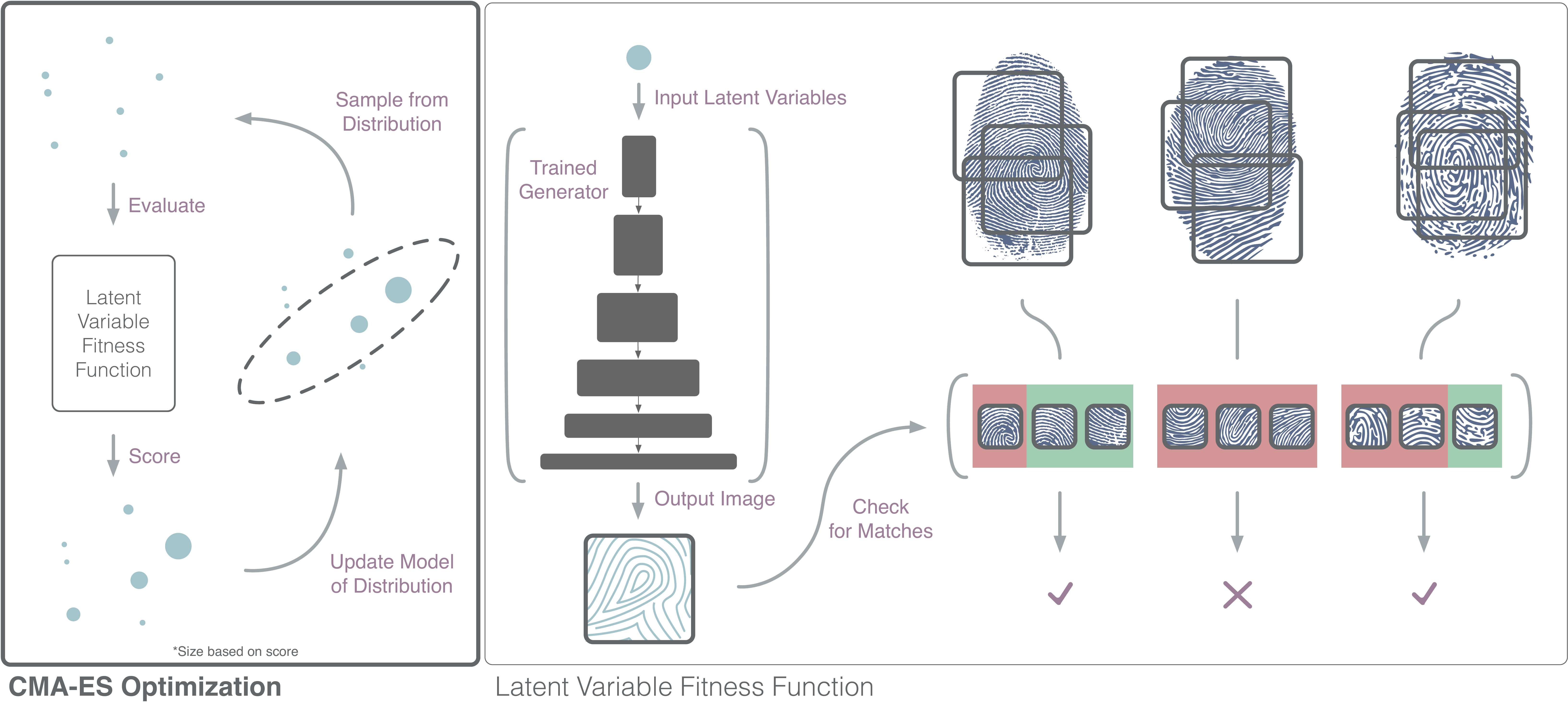}
  	\caption{Latent Variable Evolution with a trained network. On the left is a high level overview of CMA-ES and the box on the right shows how the latent variables are evaluated.}
  	\label{fig:lve}
\end{figure*}
\begin{algorithm}[tbh]
\caption{Latent Variable Evolution  \newline
$fmr \gets 1\%, .1\%, .01\%$ and $fingerprint \gets 12$ $partial$}
\label{alg:lve}
\begin{algorithmic}[1]
\STATE $G_\theta \gets \textit{trainGAN}(data)$
\FUNCTION{MatchingScore(X)}
	\STATE $img \gets G_\theta(X)$
    \STATE $score \gets 0$
	\FOR{$fingerprint$ {\bfseries in} $data$}
    	\FOR{$partial$ {\bfseries in} $fingerprint$}
			\IF{matching($img, partial, fmr$)}
            	\STATE $score++$
                \STATE \bfseries break
			\ENDIF
    	\ENDFOR
    \ENDFOR
    \STATE {\bfseries return} $score$
\ENDFUNCTION
\STATE $MasterPrint \gets \text{CMAES(MatchingScore)}$
\end{algorithmic}
\end{algorithm}


To create a DeepMasterPrint, we must evolve the latent variables of the generator to their optimal values. The inputs to a generator are called latent variables because their effect on the network output is only understood through the observed image. Since our network takes a hundred latent variables as input, the optimal solution is a point in a hundred dimensional space. As shown in Figure \ref{fig:lve}, LVE samples a number of these points, converts them to images, and then scores the images to learn the distribution over time of the best points. These optimal points are the genotypes of the DeepMasterPrints which can then be mapped to images.

LVE could use any evolutionary algorithm (or other stochastic global optimizers, such as Particle Swarm Optimization) to search the latent space. An evolutionary algorithm does not require gradients and, therefore, is ideal for black-box optimization. In this domain, the matcher can report how many identities (distinct fingerprints) match and how good each match is, but it does not provide any information about how it arrived at these results. There are no gradients that inform us which pixel of a DeepMasterPrint is most or least effective. Since the fitness score for LVE is the number of identity matches, the fitness landscape is also discontinuous. Due to the hierarchical nature of convolutional networks, the latent variables are also not independently separable. For these reasons, it is important to use an evolutionary technique that works well on rugged fitness domains, such as CMA-ES \cite{cmaes}. Since CMA-ES learns a covariance matrix of the latent variables, it can also intelligently mutate the correlated variables. In this work, we use Hansen’s Python implementation of CMA-ES \cite{cmaes2006}. To evolve each fingerprint, we let the algorithm run for 3 days.

As detailed in Algorithm \ref{alg:lve}, the fitness score is the sum total of identity matches. Each identity is represented by 12 partial fingerprints. To be verified, only one of the 12 partial fingerprint templates has to match with the input fingerprint. This is the fundamental weakness that MasterPrints and DeepMasterPrins are exploiting. The fitness of a latent variable involves converting each set of latent variables to an image, checking images against all the partial prints in the system, and then summing up the unique identities that have at least one match.

For our work, we use several different fingerprint matchers. We use the widely popular commercial fingerprint system, VeriFinger 9.0 SDK. This system is used in the fitness function in Algorithm \ref{alg:lve}. To be able to test how well the optimization for one system transfers to another, we also use the \emph{Bozorth3} matcher and the \emph{Innovatrics} IDKit 5.3 SDK. Bozorth3 is provided as an open source by NIST as part of their NIST Biometric Image Software (NBIS) suite. Both VeriFinger and Innovatrics systems can be licensed from their websites. 

\subsection{Experimental Setup}

Smartphones are the primary focus of a DeepMasterPrint attack due to their small sensors. Since smartphone systems currently use capacitive sensors, we evolve our DeepMasterPrints from a capacitive dataset using the VeriFinger matcher. To stay consistent with previous work, we evolve DeepMasterPrints for three different security levels (characterized by the False Match Rate - FMR); therefore, we get 6 DeepMasterPrints with the two generators.

In the work of Roy et al. \cite{masterprint}, they used FMRs of  1\%, 0.1\%, and 0.01\%. The FMR is the probability that an impostor (i.e., non-mate) fingerprint pair will be incorrectly marked as a match. If the FMR is set too high, the system is not very secure. If it is too low, it will reject too many genuine fingerprint pairs (i.e., mates).

To verify that our DeepMasterPrints generalize well, we split the capacitive dataset in half resulting in a test set and a training set (images in the two sets do {\em not} have any subject overlap). The test set is used for scoring the candidate DeepMasterPrints during optimization. The attack should be successful against these fingerprints as it is directly optimized for them. We test the generators on the test set to show how well the attack generalizes. 

To test the effectiveness of a DeepMasterPrint attack in the case where one does not have access to the target matcher, we test our DeepMasterPrints on two additional matchers, viz., Bozorth3 and Innovatrics. The images are neither optimized for these matchers nor the identities in the test set. Both matchers are kept as close to their default state as possible, the main parameter being the FMR. In this test scenario, the DeepMasterPrint is compared against all the identities in the test dataset to determine the number of matches.

\section{Datasets}

We model two types of fingerprint images; those scanned from inked-and-rolled impressions and those obtained from a capacitive sensor. Rolled fingerprints are produced by applying ink to the finger and rolling the finger on paper. 

\subsection{Rolled images}
The rolled fingerprints come from the publicly available NIST Special Database 9 fingerprint dataset \cite{nist9}. The dataset consists of all 10 fingerprints of 5400 unique subjects. Each fingerprint is an 8-bit grayscale image. In our work, the right thumbprint of each subject is selected. 
The images are then preprocessed by removing the whitespace and downscaling the resulting image to $256 \times 256$ pixels. To obtain partial fingerprint samples, a random $128 \times 128$ region is extracted every time an image is selected.

\subsection{Capacitive images}
The capacitive fingerprint images come from the FingerPass DB7 dataset~\cite{capacitive}. This dataset has 12 partial fingerprints for each of 720 subjects. Each partial print is of size $144 \times 144$ pixels at a resolution of 500 dpi. 
This is the same dataset that was used by Roy et al. \cite{masterprint}.

\section{Results}

\subsection{Generated fingerprints}
\begin{figure*}
  \begin{subfigure}{\linewidth}
  \includegraphics[width=.48\linewidth]{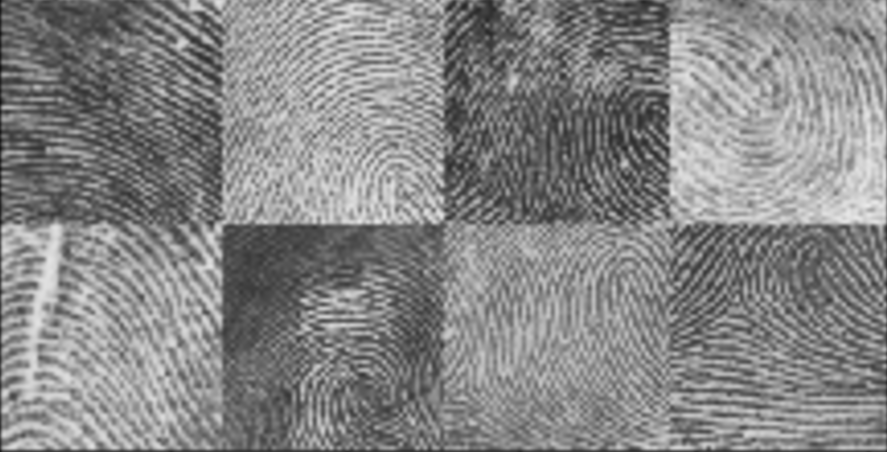}\hfill
  \includegraphics[width=.48\linewidth]{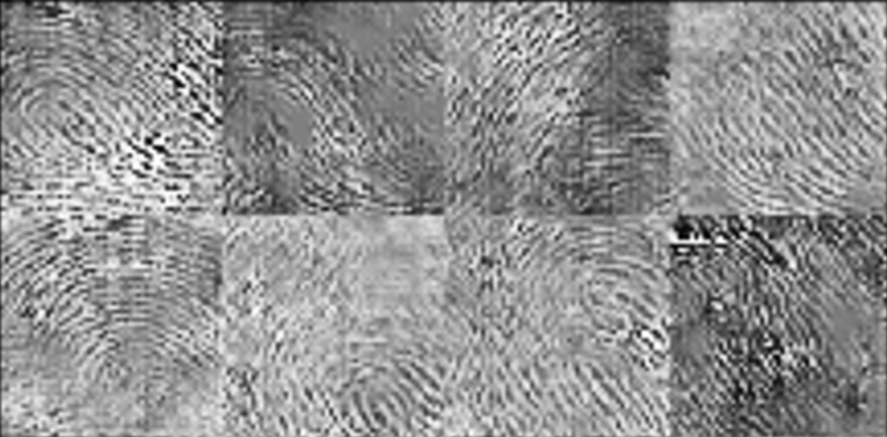}\hfill
  \caption{Real (left) and generated (right) samples for the NIST dataset.}
  \label{fig:nistGAN}
  \end{subfigure}\par\medskip
  \begin{subfigure}{\linewidth}
  \includegraphics[width=.48\linewidth]{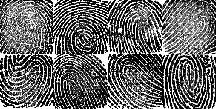}\hfill
  \includegraphics[width=.48\linewidth]{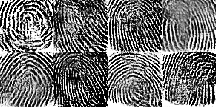}\hfill
  \caption{Real (left) and generated (right) samples for the FingerPass capacitive dataset.}
  \label{fig:capGAN}
  \end{subfigure}\par\medskip
  \caption{}
  \label{fig:gan}
\end{figure*}

The results of training the WGAN generator can be seen in Figure \ref{fig:gan}. In the right column are the generated images, while the left column contains actual samples from the datasets. The image generator seems to have captured the basic structures in both instances.

Figure \ref{fig:nistGAN} shows partial fingerprints pertaining to the rolled fingerprints from the NIST dataset. Looking at the right batch, it is clear that the generator has learned the general ridge structure of a fingerprint. Looking closer, there are certain areas that look smudged. This is most likely due to the fact that the data is generated from random sections of the fingerprint and so the generator had a difficult time learning the global shape of a full fingerprint, though it does a good job in some cases. From visual inspection, it appears to have learned the texture of fingerprints. 

Figure \ref{fig:capGAN} displays the results for the capacitive fingerprints. The results look better for this dataset. There are fewer smudges on the images and the ridges are better connected. Looking at larger batches, the generated capacitive images are  consistently better than the rolled images. 

To evaluate the images as fingerprints, we extracted the minutiae points from the image using a fingerprint matcher. The randomly generated images were determined to have similar number of minutiae points as real images in the dataset. Something interesting we noticed is that the generated images on average had double the False Match Rate as the real data. This means that even without evolution, the fingerprints are already twice as good at spoofing a system than a random real fingerprint. This suggests that the generated images display common features more often than the real data distribution. As a sanity check, we provide images of randomly generated noise to the matchers and they found no minutiae points. This means that the generator is not only producing images that look like fingerprints to humans, but they are algorithmically being identified as fingerprints too.

\subsection{DeepMasterPrints}

\begin{figure}
  \begin{subfigure}{\linewidth}
	\includegraphics[width=.3\linewidth]{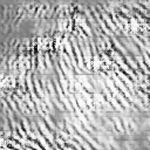}\hfill
 	\includegraphics[width=.3\linewidth]{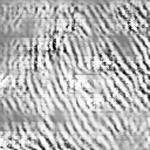}\hfill
 	\includegraphics[width=.3\linewidth]{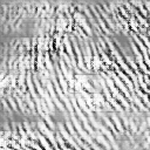}\hfill
  \end{subfigure}\par\medskip
  \begin{subfigure}{\linewidth}
 	\includegraphics[width=.3\linewidth]{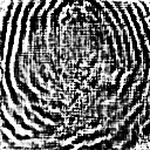}\hfill
 	\includegraphics[width=.3\linewidth]{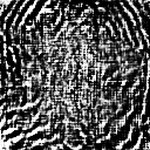}\hfill
 	\includegraphics[width=.3\linewidth]{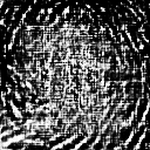}\hfill
  \end{subfigure}\par\medskip
  \caption{Evolved DeepMasterPrints for rolled fingerprints (top) and for capacitive fingerprints (bottom). Left to right, each fingerprint is optimized for an FMR of 0.01\%, 0.1\%, and 1\%, respectively.}
  \label{fig:LVEImages}
\end{figure}

The DeepMasterPrints created via LVE can be seen in Figure \ref{fig:LVEImages}. On the left are the DeepMasterPrints optimized for the higher level of security (FMR=0.01\%) and on the right are the ones for the lower level of security (FMR=1\%). The results look very similar across different security settings but not between datasets. The evolutionary algorithm is able to generate more distorted images by sampling latent variables far outside the distribution used to train the generator network. This is visually discernible in the DeepMasterPrints, with the average latent value more than three standard deviations outside the original sampling distribution in some cases. This is not necessarily a problem, as the images are still identified as fingerprints with around 20 minutiae points identified per DeepMasterPrint.

In Table \ref{tab:evolvedresults}, the percentage of false subject matches are displayed. The number of false subject matches is the number of subjects in the dataset that successfully match against the DeepMasterPrint. The second row in the table shows the results of the VeriFinger matcher when used with test data.


\begin{table*}[h]
\caption{Successful matches on the capacitive dataset. The DeepMasterPrints in Figure \ref{fig:LVEImages} are optimized for VeriFinger, at three security levels, on a capacitive training dataset. The effectiveness of each DeepMasterPrint can be seen on the test dataset.}
\label{tab:evolvedresults}
\begin{center}
\begin{tabular}{|l|c|c|c||c|c|c|}
\hline
\multirow{2}{*}{} & \multicolumn{3}{c||}{Rolled DeepMasterPrint Matches} & \multicolumn{3}{c|}{Capacitive DeepMasterPrint Matches}\\
\cline{2-7}
 & 0.01\% FMR & 0.1\% FMR & 1\% FMR & 0.01\% FMR & 0.1\% FMR & 1\% FMR\\
\hline
VeriFinger Training & 5.00\% & 13.89\% & 67.50\% & 6.94\% & 29.44\% & 89.44\%\\
\hline
VeriFinger Test & 0.28\% & 8.61\% & 78.06\% & 1.11\% & 22.50\% & 76.67\%\\
\hline
\end{tabular}
\end{center}
\end{table*}

\subsubsection{Rolled DeepMasterPrints}

The three rolled DeepMasterPrints make up the top of Figure \ref{fig:LVEImages}. At the lowest security level of 1\% FMR, a single DeepMasterPrint is able to match with 78\% of the subjects in the dataset. This is a large number of subjects, but it is unlikely that any fingerprint system uses such an FMR value. At 0.1\% FMR, the DeepMasterPrint matches 8.61\% of the dataset. This represents a much more realistic security option and results in a much higher number of (impostor) matches than what the FMR would lead one to expect. At the highest security level (FMR 0.01\%), the attack results are not very good, but this is an unlikely security level as it would be inconvenient to genuine users.


\subsubsection{Capacitive DeepMasterPrints}
The three capacitive DeepMasterPrints make up the bottom row of Figure \ref{fig:LVEImages}. Since all the match rates are for capacitive data, the capacitive DeepMasterPrints are much more visually similar to the subject data than the rolled DeepMasterPrints. This should allow the capacitive DeepMasterPrints to do better than the rolled DeepMasterPrints. Looking at Table \ref{tab:evolvedresults}, the results are, as a whole, a little better than the rolled DeepMasterPrints. At the 0.01\% FMR level, the attack results are much better. 


\subsection{Generalization}
To understand how effective this attack is, the DeepMasterPrints are tested on systems for which they have not been optimized. As stated previously, the Bozorth3 and Innovatrics matching systems are used for this purpose. The result of these tests are available in Table \ref{tab:generalization}. Both verification systems use the same three FMRs used for VeriFinger.

In the case of VeriFinger, six different DeepMasterPrints were used. This represents the case where the target system is known and can be accessed or replicated to launch a more highly optimized attack. In these cases, we found better performance by optimizing for each security level. This strategy did not prove very effective for the case where the test environment is unknown. It was found that evolving DeepMasterPrints at high security settings generalized the best. Therefore, the results reported  are based on the two DeepMasterPrints that were optimized for an FMR of 0.01\%  and used against the two verification systems at all three security settings. 

Bozorth3 is publicly available and free to use, but it is also an older matcher. Perhaps this explains why the Rolled DeepMasterPrint generalize so well to this matcher. The DeepMasterPrint actually does better with Bozorth3 than with VeriFinger in this case. 
The capacitive fingerprint was much less effective against this system but still successful overall. At all security levels, except at the highest one, the DeepMasterPrint performs around 30 times better than an average fingerprint. At 0.01\% FMR there are no matches; this makes it difficult to accurately determine the DeepMasterPrint performance at this level.

Innovatrics is a more recent matcher still under active development. It would, therefore, be expected to be more resilient to attacks using DeepMasterPrints. Surprisingly, the capacitive DeepMasterPrint is consistent here and gets similar results to what it did on Bozorth3. One hypothesis here is that the capacitive DeepMasterPrint has found some universal patterns that are not specific to a particular verification system. The rolled DeepMasterPrint actually does worse than the capacitive one in spite of performing so well on Bozorth3. It is evident that these two matchers handle rolled fingerprints very differently. The training data used to train the fingerprint generator definitely makes a difference here. The DeepMasterPrints are roughly 10 times more effective than a random image.

\begin{table}[h]
\caption{The DeepMasterPrints optimized for the highest security levels were found to generalize the best. The two DeepMasterPrints optimized for 1\% FMR are tested on the Bozorth3 and Innovatrics matchers. They are both tested at three different security levels, with the percentage of successful matches on the capacitive test set reported.}
\label{tab:generalization}
\begin{center}
\begin{tabular}{|l r|r|r|}
\hline
\multicolumn{2}{|c|}{Verification System} & \multicolumn{2}{c|}{MasterPrint Matches}\\
\hline
 & FMR & Rolled & Capacitive\\
\hline
\multirow{3}{*}{Bozorth3} & 0.01\% & 0.00\% & 0.00\%\\
\cline{2-4}
 & 0.1\% & 23.06\% & 2.78\%\\
\cline{2-4}
 & 1\% & 89.72\% & 31.39\%\\
\hline
\multirow{3}{*}{Innovatrics} & 0.01\% & 0.00\% & 0.83\%\\
\cline{2-4}
 & 0.1\% & 0.83\% & 3.61\%\\
\cline{2-4}
 & 1\% & 10.56\% & 25.28\%\\
\hline
\end{tabular}
\end{center}
\end{table}


\subsection{Comparative Results}

In our work, we created a DeepMasterPrint that is intended to spoof an arbitrary identity in a single try. Previous work had much worse results when given only a single attempt. Besides providing an image, LVE creates a much more effective MasterPrint. Table \ref{tab:compareresults} has the results of the minutiae-only approaches and the capacitive DeepMasterPrint image \cite{compareData}. 
In the previous work by Roy et al. \cite{masterprint}, the authors generated a suite of five fingerprint templates that were used sequentially to launch an attack, assuming five attempts. Our results for a single DeepMasterPrint is comparable to this suite of multiple MasterPrints. We expect LVE to do very well in creating sequential DeepMasterPrints.

\begin{table}[h]
\caption{Percentage of subjects matched using the DeepMasterPrint compared to the previous method for generating MasterPrints. The results are on the capacitive dataset and uses the VeriFinger matcher.}
\label{tab:compareresults}
\begin{center}
\begin{tabular}{|l|c|c|c|}
\hline
 & \small 0.01\% FMR & \small 0.1\% FMR & \small 1\% FMR \\
\hline
Single MasterPrint & 1.88\% & 6.60\% & 33.40\% \\
\hline
MasterPrint Suite & 6.88\% & 30.69\% & 77.92\% \\
\hline
Single DeepMasterPrint & 1.11\% & 22.50\% & 76.67\% \\
\hline
\end{tabular}
\end{center}
\end{table}

\balance 

\section{Conclusion}

This paper presents Latent Variable Evolution as a method for generating DeepMasterPrints: partial fingerprint images which can be used for launching dictionary attacks against a fingerprint verification system. The first step is to train a GAN using images from a fingerprint dataset. Then LVE searches the latent variables of the generator network for an image that maximizes the number of fingerprints which are successfully matched with it. The method proposed in this paper was found to (1) result in DeepMasterPrints that are more successful in matching against fingerprints pertaining to a large number of distinct identities, and (2) generate complete images - as opposed to just minutiae templates - which can potentially be used to launch a practical DeepMasterPrint attack. Experiments with three different fingerprint matchers and two different datasets show that the method is robust and not dependent on the artifacts of any particular fingerprint matcher or dataset.


Beyond the application of generating DeepMasterPrints, this paper successfully shows the usefulness of searching the latent space of a generator network for images, or other artifacts, that meet a given objective. This idea is surprisingly under-explored and could be useful in computational creativity research as well as other security domains. Initial work on using a similar approach for aesthetic purposes in an interactive setting can be found in~\cite{deepIE}.

{\small \em
}

{\small 
\bibliographystyle{ieee}
\bibliography{bibliography}
}
\end{document}